\documentclass[11pt]{article}

\usepackage[margin=1in]{geometry}
\usepackage{times}
\usepackage[T1]{fontenc}
\usepackage{booktabs}
\usepackage{array}
\usepackage{longtable}
\usepackage{hyperref}
\usepackage{xcolor}
\usepackage{enumitem}
\usepackage{amsmath}
\usepackage{amssymb}
\usepackage{natbib}
\usepackage{comment}
\usepackage{graphicx}
\hypersetup{
    colorlinks=true,
    linkcolor=blue!50!black,
    urlcolor=blue!50!black,
    citecolor=blue!50!black,
    pdftitle={FACTS-MAS: A Multi-Agent Architecture for Adaptive Forecasting from Heterogeneous Evidence},
    pdfauthor={Madumita Karthikeyan, Umang Rajnikant Bid, Rong Pan}
}

\title{\textbf{FACTS-MAS: A Multi-Agent Architecture for\\
Adaptive Forecasting from Heterogeneous Evidence}}

\author{
Madumita Karthikeyan\\
\texttt{mkarthi5@asu.edu}
\and
Umang Rajnikant Bid\\
\texttt{umangraj@asu.edu}
\and
Rong Pan\\
\texttt{rong.pan@asu.edu}
}

\begin{document}

\maketitle

\begin{abstract}
\noindent
Real-world forecasting often requires integrating heterogeneous evidence whose predictive value varies across horizons, operating regimes, and domains. We introduce FACTS-MAS, a domain-agnostic multi-agent forecasting framework that decomposes evidence into specialized channels for temporal dynamics, exogenous context, seasonality, structural information, and discrete events. Each agent generates an independently testable forecast through a common interface, while leakage-safe historical validation measures its predictive skill relative to a naive reference. These skill estimates are used to weight and combine agent forecasts, so that uninformative evidence is either removed or downweighted rather than forced to participate. A calibration procedure selects domain-specific constant means based on historical validation folds. We evaluate FACTS-MAS on weekly for-sale housing inventory from Zillow and weekly unemployment insurance claims from FRED, for 15 U.S.\ metropolitan areas at 4-, 8-, and 13-week forecasting horizons. Our method outperforms both a naive last-value forecast method and the TimesFM foundation model. However, a factor-adjusted exponential smoothing model performs as well as the fused forecast. The evidence therefore supports FACTS-MAS as a systematic protocol for testing, calibrating, and recomposing evidence channels, before assembling them to  the strongest single model for a specific application domain. Code, data, and results are available at \url{https://github.com/Madumita24/FACTS-MAS}.

\end{abstract}

\section{Introduction}
\label{sec:intro}

Forecasting in operational settings almost always draws on more than one
kind of information. A target series may be strongly autoregressive, yet its
near-term path can also depend on exogenous conditions, recurring seasonal
cycles, slowly varying attributes of the entity being forecast, and discrete
events that do not arrive as ordinary numerical features. These information sources
differ in representation, timing, and, most importantly, predictive value.
A signal that helps at a short horizon or in one regime may add nothing at a
longer horizon or in another domain.

A common approach in forecasting is to fold all available inputs into a single
statistical/machine-learning (ML) model or a single artificial intelligence (AI) reasoning process. This forecasting approach
may be able to capture complex interactions among different information sources,
but it binds three tasks into one object: how each source is encoded, how the
forecast is produced, and how credit is assigned to each source.
When the forecast accuracy has changed, it is difficult to identify whether the change came from
temporal dynamics, external context, structure, or events. It also requires dissimilar information sources to be
translated into a single input format, even though time-series dynamics,
calendar effects, static covariates, and unstructured events are not naturally
expressed in the same way.

In this paper, we develop a new approach, Forecasting from Adaptive,
Calibrated, Testable Sources -- Multi-agent AI System (FACTS-MAS), which is
built upon the opposite premise. We define AI agents by their
\emph{functional evidence roles}, not by an application name. Each agent
turns the evidence in its channel into a numerical forecast that satisfies a
common output contract. The fusion step then asks a focused question: \emph{does this
channel improve on a simple reference forecast in leakage-safe historical
validation?} We quantify the relative skill of each agent for the
targeted forecasting task as a weight; weak or redundant channels are
downweighted rather than forced to participate. Domain-specific constants like event-impact scales, event durations, and
reversion caps are chosen by the same validation test, and these constants
can be set to zero. This changes the interpretation of negative results. If an agent receives
little weight, it is not necessarily a failed component; instead, it is
evidence that the channel it is managing contains little incremental
information for the current domain, or that the present implementation does
not extract the useful information. Likewise, a plausible causal event may in fact have little
predictive power in one domain, but we can keep it and assign it a
confidence score for later domains.

We evaluate FACTS-MAS on the forecast of weekly U.S.\ housing inventory, a setting in
which all five candidate channels can be populated: recent inventory
dynamics, national financial conditions, annual listing seasonality,
metropolitan structure, and disaster declarations. The study covers 15
metropolitan statistical areas (MSAs) at forecast horizons of 4, 8, and 13
weeks. At each forecast origin, defined as the week in which a prediction is issued,
models are trained only on observations dated on or before that week; because
Zillow publishes each week's value one to two weeks later, we also report every result with a two-week publication lag. Evaluation uses a six-fold expanding-window
backtest, with a 13-week embargo between the end of training and the start
of testing, followed by a single pre-registered holdout on data after
February 21, 2026 that was not accessed until the configuration was frozen.
On Zillow's smoothed inventory series, the frozen configuration reduces MAPE
relative to a naive last-value forecast from 4.194\% to 2.118\% at the 4-week
horizon, from 7.225\% to 3.622\% at 8 weeks, and from 10.814\% to 5.731\% at
13 weeks in the backtest, and from 3.122\% to 1.243\%, 5.445\% to 1.929\%, and
8.438\% to 3.045\% on the holdout. The reduction is smaller on the unsmoothed
series (24\% at 4 weeks in the backtest, against 50\% on the smoothed series).
The fused forecast is significantly more accurate than the TimesFM foundation
model \citep{das2024timesfm} in the backtest. On the holdout it has a lower
error than a chain-of-thought language-model forecast built on the NEXUS
prompt \citep{das2026nexus}, but the difference is not statistically
significant. It is not more accurate than a
factor-adjusted damped exponential smoothing model, with which it is
statistically equivalent at 8 weeks. We also transfer the method to weekly
unemployment insurance claims to assess the generalizability of the approach.

The contributions of this paper are as follows.

\begin{enumerate}[leftmargin=*]
    \item We introduce FACTS-MAS, a multi-agent forecasting procedure that
    assigns heterogeneous evidence to functional channels---temporal,
    exogenous, seasonal, structural, and event-based---and requires each
    agent to return a numerical forecast under a common output contract.

    \item We develop a leakage-safe fusion rule that measures each agent's
    improvement over a naive reference, aggregates that skill by the median,
    and converts the resulting scores into fusion weights. The weights may
    depend on forecast horizon and operating regime when the validation
    sample is sufficient to support that conditioning; a regime is used only
    when it contains at least 13 distinct validation weeks, and otherwise the
    pooled weights apply.

    \item We specify a calibration procedure for domain-specific constants,
    including event-impact scales, event durations, and reversion caps. Candidate values are
    selected on historical validation folds, and a constant may be set to
    zero when validation does not support a nonzero effect.

    \item We evaluate the method on weekly for-sale housing inventory for
    15 U.S.\ metropolitan areas at forecast horizons of 4, 8, and 13 weeks,
    on both Zillow's smoothed and unsmoothed series. The analysis compares
    accuracy with naive, seasonal-naive, and seasonally adjusted naive
    references, with exponential-smoothing and Theta models, with the
    TimesFM foundation model, and with a chain-of-thought language-model
    forecast, using panel Diebold--Mariano tests and block-bootstrap
    confidence intervals. It distinguishes evidence channels that add
    incremental skill from channels whose quantitative effect is weak or
    unsupported.

    \item We report a pre-registered evaluation: the configuration, the
    hypotheses, and the statistical tests were fixed and committed before a
    holdout period of 29 weekly origins was accessed, and every result on
    that holdout is reported, including those that do not favor the fused
    forecast.
\end{enumerate}

The remainder of the paper is organized as follows.
Section~\ref{sec:related} reviews multi-agent forecasting, the evolution of
forecasting models, forecast combination, and the NEXUS precursor.
Section~\ref{sec:framework} defines the FACTS-MAS procedure, including the
evidence channels, the shared forecast interface, skill-based fusion,
calibration, and attribution.
Section~\ref{sec:methodology} states the rolling-origin protocol, the
information boundary, the accuracy metrics, and the comparison design, then
instantiates the method for U.S.\ housing inventory and reports forecast
accuracy together with the channel and fusion diagnostics.
Section~\ref{sec:discussion} discusses the housing findings, including the
holdout results, and the transfer of the method to weekly unemployment
insurance claims.
Section~\ref{sec:conclusion} concludes.

\section{Related Work}
\label{sec:related}

FACTS-MAS draws on four lines of research: multi-agent forecasting,
the succession of time-series model classes, the combination of
heterogeneous evidence, and LLM-assisted interpretation of unstructured
information. This section reviews each line and states how FACTS-MAS differs.

\subsection{Multi-Agent and LLM-Assisted Forecasting}

Multi-agent systems decompose a complex prediction task into specialized
components whose outputs are orchestrated or aggregated
\citep{chang2025tsr,wang2024agents}. Large language
models have widened this pattern: agents can interpret text, call tools,
plan, and communicate over mixed structured and unstructured inputs
\citep{yao2023react,jang2026wit}.
Recent agentic forecasting systems organize those capabilities into
workflows---planning, tool use, iterative revision, and ensemble
generation---rather than a single-pass map from history to the future
\citep{tao2026castr1,pan2026castflow}.
Role-specialized designs give agents distinct duties and pass intermediate
results through conversation or a shared workflow
\citep{park2023generative,wu2024autogen,hong2024metagpt}.

However, specialization alone does not establish that every agent adds predictive
information. Two agents may encode the same signal, a plausible feature
family may have no out-of-sample benefit, and usefulness may depend on the
forecast horizon or the operating regime. When agents only converse, their
individual numerical contributions cannot be isolated. FACTS-MAS differs by requiring each agent to issue its own numerical
forecast and by measuring that forecast's skill relative to a simple
reference before fusion assigns it any weight. The skill measurement may
depend on the forecast horizon and the operating regime when the validation
sample is sufficient.

\subsection{From Statistical Models to Foundation Models}

Time-series forecasting has moved through several model classes that remain
useful as channel implementations or as baselines. Exponential smoothing,
including the Holt--Winters trend form, forecasts from a local level and
trend
\citep{holt2004exponential,winters1960forecasting,hyndman2002statespace,pan2011holtwinters}.
Additive statistical models of the same family are widely used in practice
\citep{taylor2018prophet}.
Large competitions then compared these statistical methods with later
machine-learning entries on common collections of series
\citep{makridakis2000m3,makridakis2020m4}.
The M4 winner combined exponential smoothing with a recurrent network
\citep{smyl2020hybrid}.
Global neural forecasters fit one model to many related series, including
autoregressive recurrent networks \citep{salinas2020deepar}, residual basis
expansions \citep{oreshkin2020nbeats}, and attention models that also accept
covariates \citep{lim2021tft}.
\citet{lim2021survey} survey this deep-learning line.

Transformer forecasters adapted attention to long horizons
\citep{zhou2021informer,wu2021autoformer,nie2023patchtst}, and
\citet{wen2023transformers} survey that design family.
Subsequent comparisons showed that simple linear models can match or beat
several of those architectures \citep{zeng2023transformers}.
Pretrained foundation models now produce zero-shot forecasts across domains,
including TimesFM \citep{das2024timesfm}, Chronos \citep{ansari2024chronos},
and Moirai \citep{woo2024moirai}.
Those zero-shot forecasts can still carry variance and systematic bias.
Bootstrap aggregation, regression stacking, and residual correction of a
foundation-model forecast have reduced both on an operational electricity-load
series \citep{modi2025tsfm}.
A parallel line asks a language model to forecast, either by reprogramming a
frozen model \citep{jin2024timellm} or by encoding the series as text
\citep{gruver2023llmtime}.
FACTS-MAS does not replace this sequence of forecasters. A channel may use a
statistical smoother, a regression, or a pretrained model. The housing study uses Holt's linear-trend exponential smoothing on the
temporal channel at the 4-week horizon and a factor-adjusted damped-trend
model \citep{gardner1985damped} at 8 and 13 weeks, and it uses TimesFM~2.5,
a later checkpoint of the TimesFM line, as a numerical baseline.

\subsection{Heterogeneous Evidence and Forecast Combination}

Operational forecasting problems commonly pool evidence with heterogeneous
statistical structure: endogenous dynamics, exogenous covariates, calendar
effects, slowly varying entity attributes, and discrete events.
Recurrent events can be forecast through the cumulative intensity of the
event process, including with boosted trees that use both static and dynamic
features \citep{liu2021boostr}. A joint
statistical or machine-learning model maps this collection to a single
forecast, so that input encoding, prediction, and source-level attribution
are determined by one fitted mapping. Forecast combination retains a separate
forecast for each source and aggregates those forecasts.
Equal-weight aggregation reduces dependence on any one specification
\citep{clemen1989combining}.
Performance-based weights assign greater influence to component forecasts
with lower historical error
\citep{bates1969combination}.
Regression on the component forecasts is another classical route to those
weights \citep{granger1984combining,stock2004combination}.
Static weights assume that relative accuracy is sufficiently stable to be
pooled across time \citep{newbold1974experience,wang2023combinations}.
\citet{timmermann2006combinations} reviews when such combination improves on
the best individual forecast. That question is central here: a combination of channels is only useful if
it improves on the strongest single model that could have been selected on
the same validation data.
A related prediction task is degradation under a time-varying environment,
where an ensemble is trained to follow that change
\citep{wang2020degradation}.

FACTS-MAS uses rolling weights based on each agent's skill relative to a
simple reference forecast. The skill
score, its median aggregation, and the softmax conversion are defined in
Section~\ref{sec:framework}. Whether those weights should also depend on
forecast horizon or operating regime is treated there as a validation
choice, because finer conditioning is estimated from fewer observations.

\subsection{Calibration and NEXUS}

Plausible numerical assumptions---event-severity maps, effect durations,
lags, and thresholds---can degrade forecast accuracy even when the underlying
causal mechanism is plausible. FACTS-MAS selects such constants on historical
validation folds and allows a constant to be set to zero when a nonzero
effect is unsupported. The calibration rule is given in
Section~\ref{sec:framework}.

The immediate precursor of our MAS approach is NEXUS \citep{das2026nexus}, which forecasts from
mixed structured and unstructured evidence by passing that evidence through
multi-agent language-model reasoning and then issuing the forecast.
Efficiency extensions of that design, such as early-exit gating and batched
reasoning, retain language-model reasoning on the main forecasting path; they
are not part of FACTS-MAS and are not evaluated here. FACTS-MAS assigns the
same evidence to specialized channels and obtains an independent numerical
forecast from each channel. The fusion weights are estimated from the
channels' historical skill relative to a simple reference on a validation
window that ends at the training cutoff, and they are re-estimated whenever
the training window is extended. Those weights may also depend on the
forecast horizon and the operating regime when the validation sample is
sufficient.

The housing experiments in Section~\ref{sec:housing} compare FACTS-MAS with
a chain-of-thought language-model forecast that uses the NEXUS prompt, with a
time-series foundation model, and with strong statistical references.
Those comparisons test the forecasting consequence of this decomposition.

\section{FACTS-MAS: Domain-Agnostic Forecasting Architecture}
\label{sec:framework}

FACTS-MAS separates a reusable forecasting procedure from any one
domain's data and models. The appropriate raw variables, predictive models,
event types, and operating regimes may depend on the domain, but the
forecasting procedure remains the same. Heterogeneous evidence is digested
and fed into forecasting channels; each channel produces an independently
evaluable forecast, and these forecasts are combined adaptively.

Let $y_t$ denote the target variable observed up to forecast origin $t$, and
let $H$ denote the requested forecast horizon. Let $\mathcal{I}_t$ denote the
information available at that origin. For a forecasting domain
$\mathcal{D}$, this information is streamed into $K$ functional
evidence channels,

\begin{equation}
    \mathcal{E}^{(\mathcal{D})}
    =
    \left\{
    E_1^{(\mathcal{D})},
    E_2^{(\mathcal{D})},
    \ldots,
    E_K^{(\mathcal{D})}
    \right\}.
\end{equation}
$E_k^{(\mathcal{D})}$ denotes the channel, and each channel is assigned one
agent. The assigned agent reads that channel and maps the available
information to a horizon forecast,

\begin{equation}
    f_k : \mathcal{I}_t \rightarrow \hat{\mathbf{y}}_k,
    \qquad k=1,\ldots,K,
\end{equation}
where $\hat{\mathbf{y}}_k$ is the forecast over the requested horizon.
In a named domain $\mathcal{D}$, the agent is $A_k$,

\begin{equation}
    A_k:
    \mathcal{I}_t^{(\mathcal{D})}
    \rightarrow
    \hat{\mathbf{y}}_{k,t}^{(H)},
\end{equation}
where $\hat{\mathbf{y}}_{k,t}^{(H)}
=
\left[
\hat{y}_{k,t+1},
\ldots,
\hat{y}_{k,t+H}
\right]$
is the $H$-step forecast produced independently by agent $k$.

These maps determine the forecast. At each forecast origin, domain evidence is
assigned to the $K$ channels, and each agent returns an independent forecast.
A validation step scores each agent's historical forecasts against a
reference, the resulting skill scores set the fusion weights, and adaptive
fusion combines the agent forecasts into $\hat{\mathbf{y}}_{\mathrm{FACTS}}$.
Calibration is a separate, offline step that selects channel constants on
historical validation folds before the test period begins.

\begin{figure}[t]
\centering
\includegraphics[width=\textwidth]{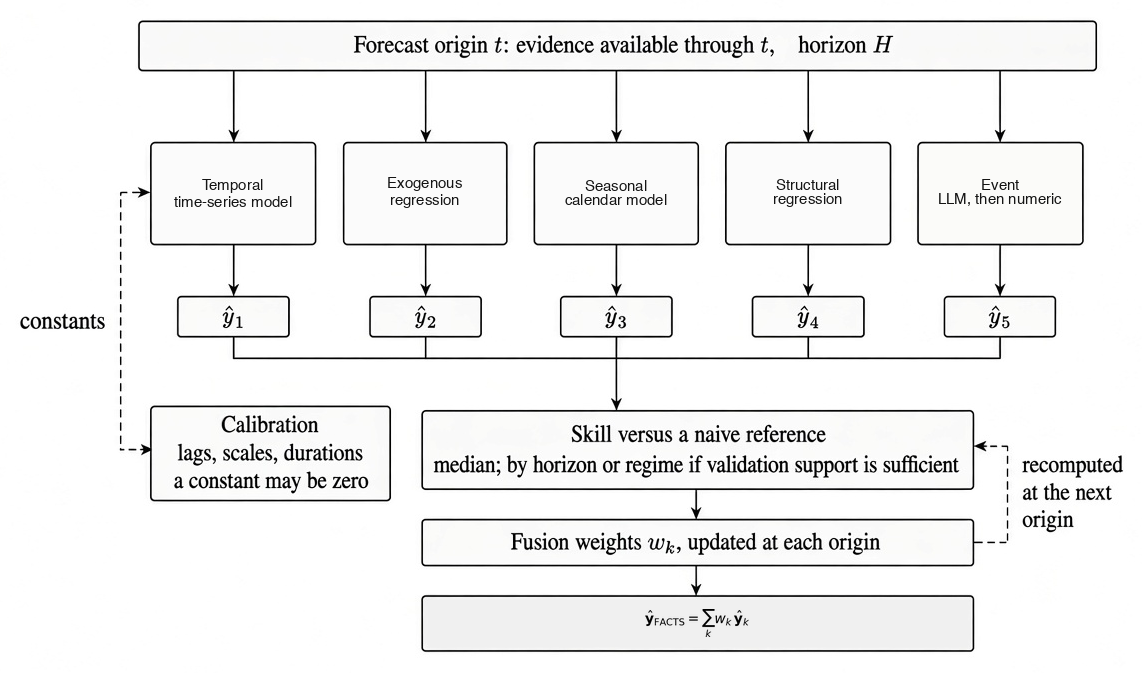}
\caption{FACTS-MAS at one forecast origin. Each evidence channel returns an
independent horizon forecast. Skill relative to a naive reference, measured on
a validation window that ends at the training cutoff, sets the fusion weights.
Calibration acts offline on channel constants and may set a constant to zero.}
\label{fig:architecture}
\end{figure}

Figure~\ref{fig:architecture} shows this procedure at one forecast origin.
The following subsections describe the components of this process and identify
which elements remain invariant across domains and which need to be
reconfigured for specific forecasting tasks.

\subsection{Functional Evidence-Channel Decomposition}
\label{sec:evidence_channels}

FACTS-MAS organizes forecasting evidence according to its functional role
rather than according to a particular dataset or application. The current
architecture defines five candidate evidence channels:

\begin{enumerate}[leftmargin=*]

    \item \textbf{Temporal channel.}
    It models the endogenous dynamics of the target using its historical
    trajectory. This channel captures stability, trend, autocorrelation,
    and other temporal structure that can be estimated without external
    information.

    \item \textbf{Exogenous channel.}
    It represents external numerical variables that may influence future
    changes in the target. The identity and transmission lag of these
    variables are domain dependent.

    \item \textbf{Seasonal channel.}
    It represents recurring temporal structure associated with calendar,
    periodic, or cyclic behavior.

    \item \textbf{Intrinsic or structural channel.}
    It represents relatively persistent characteristics of the entity being
    forecast. These features may distinguish entities even when they share
    similar recent temporal trajectories.

    \item \textbf{Event channel.}
    It represents discrete or unstructured information whose forecasting
    relevance may require semantic interpretation before it can be converted
    into a structured signal.

\end{enumerate}

These channels are \emph{functional abstractions}, not fixed feature sets.
For example, the exogenous channel may contain monetary-policy indicators in
a housing application, weather and demand drivers in an energy application,
or labor-market indicators in an employment application. Similarly, the
meaning of a relevant event depends on the target domain.

This distinction is central to domain transfer. FACTS-MAS does not assume
$E_k^{(\mathcal{D}_1)} = E_k^{(\mathcal{D}_2)}$.
Instead, it assumes that both domains can expose evidence serving comparable
forecasting functions, $E_k^{(\mathcal{D}_1)} \sim E_k^{(\mathcal{D}_2)}$
with respect to functional role, while allowing their data sources, features,
transformations, and agent implementations to differ.

\subsection{Shared Agent Interface}
\label{sec:agent_interface}

To keep evidence-channel implementation separate from downstream fusion,
every forecasting agent must satisfy a common output contract. For a given
forecast request, an agent returns

\begin{equation}
O_k =
\left(
a_k,
d,
t,
H,
\hat{\mathbf{y}}_k,
\mathbf{c}_k
\right),
\end{equation}
where $a_k$ identifies the agent, $d$ identifies the entity or forecasting
unit, $t$ is the forecast origin, $H$ is the requested horizon,
$\hat{\mathbf{y}}_k$ contains exactly $H$ forecast values, and
$\mathbf{c}_k$ optionally contains horizon-specific confidence values.

This output contract enforces three important invariants. First, all agents must
produce forecasts for the same entity, origin, and requested horizon before
fusion. Second, the output length must exactly match the requested horizon.
Third, target forecasts must satisfy domain-valid output constraints; for the
housing implementation, for example, inventory forecasts are constrained to
be non-negative.

The shared contract allows an agent implementation to be replaceable without
modifying the fusion layer. A statistical model, machine-learning model,
foundation time-series model, or LLM-assisted semantic embedding model can therefore occupy
an evidence channel as long as its output satisfies the same forecasting
interface.

\subsection{Leakage-Safe Agent Evaluation}
\label{sec:agent_evaluation}

The quality of a forecasting agent is estimated using its historical forecasts
generated under the same information constraints that would exist during
deployment. For each
forecast origin $t$, an agent may access only evidence available on or before
that origin after accounting for publication and availability delays.

For an agent $k$ and reference model $b$, define the absolute percentage
errors at validation point $i$ as

\begin{equation}
e_{k,i}
=
\frac{
\left|y_i-\hat{y}_{k,i}\right|
}{
\left|y_i\right|
},
\qquad
e_{b,i}
=
\frac{
\left|y_i-\hat{y}_{b,i}\right|
}{
\left|y_i\right|
}.
\end{equation}
FACTS-MAS measures incremental forecasting skill relative to the reference,
$s_{k,i}=1-e_{k,i}/e_{b,i}$.
Thus,

\begin{equation}
s_{k,i}
\begin{cases}
>0, & \text{agent improves on the reference},\\
\approx0, & \text{agent approximately matches the reference},\\
<0, & \text{agent underperforms the reference}.
\end{cases}
\end{equation}
Skill is scored against the persistence baseline, so an agent that only
reproduces that baseline receives a skill near zero.

Directly averaging the pointwise skill ratios can be unstable when
the reference error is extremely small at individual validation points.
FACTS-MAS therefore aggregates pointwise skill using the median,

\begin{equation}
\tilde{s}_k
=
\operatorname{median}_{i \in \mathcal{V}}
\left(s_{k,i}\right),
\label{eq:median_skill}
\end{equation}
where $\mathcal{V}$ denotes the leakage-safe validation window. The median
reduces the influence of isolated points at which an unusually small
reference error would otherwise produce an extreme ratio.

\subsection{Skill-Based Adaptive Fusion}
\label{sec:fusion}

The validated skill scores are converted into non-negative
normalized weights using a temperature-controlled softmax,

\begin{equation}
w_k
=
\frac{
\exp(\tilde{s}_k/\tau)
}{
\sum_{j=1}^{K}
\exp(\tilde{s}_j/\tau)
},
\label{eq:softmax}
\end{equation}
where $\tau>0$ controls how strongly differences in historical skill affect
the final allocation. In the housing configuration $\tau=0.05$. This value
was set by hand from the magnitude of validation skill scores, not selected by
a search; a validation-selected temperature was tested and not adopted
(Section~\ref{sec:fusion_ablation}).

The fused forecast is then

\begin{equation}
\hat{\mathbf{y}}_{\mathrm{FACTS},t}^{(H)}
=
\sum_{k=1}^{K}
w_{k,t}^{(H)}
\hat{\mathbf{y}}_{k,t}^{(H)}.
\label{eq:fusion}
\end{equation}

Softmax weighting is useful in this setting because it preserves the complete
ranking of agents, including agents with negative relative skill. No hard
threshold is required to assign identical minimum weights to all
underperforming channels. An agent that is slightly worse than the reference
can therefore remain distinguishable from an agent that is substantially
worse.

Weights are re-estimated whenever the training window is extended rather than
fixed once for the entire evaluation period. In the backtest, each fold
estimates its weights from a 104-week validation window ending at the training
cutoff and holds them fixed over that fold's test window. Consequently,
$w_{k,t_1}^{(H)} \neq w_{k,t_2}^{(H)}$ is permitted across folds, and within
a fold when $t_1$ and $t_2$ fall in different regimes.

\subsection{Adaptive Context and Regime Conditioning}
\label{sec:adaptive_context}

The relative value of an evidence channel may depend on both forecast horizon
and operating context. FACTS-MAS therefore supports multiple levels of
conditioning.

Let $r_t$ denote a domain-specific regime observed at forecast origin $t$.
The regime variable itself is not part of the invariant framework. It is a
domain-specific configuration satisfying two requirements: it must represent
a meaningful state for the target process, and its value at forecast origin
$t$ must be computable without future information. The same historical skill
scores can be grouped in four ways. Pooled weighting, $w_k$, uses the full
validation history and assigns each agent one weight at every horizon and in
every regime. Horizon-conditioned weighting, $w_k^{(H)}$, recomputes skill
at each forecast horizon. Regime-conditioned weighting, $w_k^{(r)}$,
recomputes skill within each regime. Horizon--regime-conditioned weighting,
$w_k^{(H,r)}$, recomputes skill inside each combination of horizon and
regime. FACTS-MAS treats the choice among these groupings as a validation
problem: for each eligible configuration, competing fusion strategies are
evaluated on historical validation data, and the strategy with lower
validation error is selected before entering the test window.

This creates an important distinction between \emph{adaptivity} and
\emph{granularity}. A more granular rule can represent more context-specific
behavior but is estimated from fewer observations. Rare regimes can therefore
produce insufficient evidence for reliable weight selection. FACTS-MAS
allows the level of adaptation to be constrained by historical support rather
than assuming that finer conditioning always improves generalization.
In the housing configuration, a regime-specific weight vector is used only
when the regime contains at least 13 distinct validation weeks; otherwise the
forecast falls back to the pooled weights. An earlier version of the
implementation had no such rule and assigned equal weights to every channel
when a regime had no validation data, which affected 780 test cells and
inflated the apparent benefit of regime conditioning.

\subsection{Calibration of Domain-Specific Parameters}
\label{sec:calibration}

Some evidence channels require domain-specific constants or mappings. Examples
include event-effect duration, numerical impact scales, classification
thresholds, lag lengths, or transformation parameters. Assigning such values
from intuition alone can introduce an unmeasured source of forecasting error.

FACTS-MAS therefore separates parameter calibration from held-out evaluation.
For a domain-specific parameter $\theta$ and candidate set
$\Theta=\{\theta_1,\ldots,\theta_m\}$, each candidate is evaluated on
historical validation data for validation fold $f$,

\begin{equation}
L_f(\theta)
=
\operatorname{MAPE}
\left(
\mathbf{y}_f,
\hat{\mathbf{y}}_f(\theta)
\right).
\end{equation}

A candidate is not promoted solely because it achieves the lowest pooled
error. The framework additionally checks whether the direction of improvement
is sufficiently consistent across folds. This discourages parameter changes
whose apparent benefit is driven by a small subset of historical periods.
Most importantly, calibration is permitted to return a null quantitative
effect. If $\theta^{*}=0$
provides the best and most consistent validation performance, the framework
interprets this as evidence that the corresponding magnitude adjustment is
unsupported. Other outputs from the same channel, such as classification,
confidence, or attribution information, can remain available even when its
direct numerical adjustment is removed.

In the housing study, each constant was selected once using the validation
windows of all six folds, with a candidate promoted only if it improved
validation loss in at least five of the six folds. Because later folds'
validation windows postdate earlier folds' test windows, this selection is not
fully nested in time. The only adopted calibration result, a zero event-impact
scale, changes test accuracy by at most 0.002 percentage points.

\subsection{Factor Attribution}
\label{sec:attribution}

Because every agent produces an independent forecast and every fusion weight
is derived from historical predictive skill, FACTS-MAS exposes a direct
forecast-level attribution mechanism.

For forecast origin $t$, the contribution of agent $k$ to the forecasted
change from the last observed value is

\begin{equation}
C_{k,t}^{(H)}
=
w_{k,t}^{(H)}
\left(\hat{\mathbf{y}}_{k,t}^{(H)} - y_t\mathbf{1}\right),
\end{equation}
so that $\sum_k C_{k,t}^{(H)} = \hat{\mathbf{y}}_{\mathrm{FACTS},t}^{(H)} - y_t\mathbf{1}$.
A level-based share $w_{k,t}^{(H)}\hat{\mathbf{y}}_{k,t}^{(H)}$ would credit
an agent that merely repeats the last observed value with a share of the
forecast level even though it contributes nothing to the forecasted change;
the change-based form assigns such an agent a contribution of zero.
The set
$\left\{w_{1,t}^{(H)},\ldots,w_{K,t}^{(H)}\right\}$
provides an interpretable summary of which evidence channels were historically
most useful under the validation conditions governing that forecast.

These weights should not be interpreted as causal effect estimates. They are
empirical measures of relative predictive contribution under the selected
forecasting and validation procedure. The distinction is important: an agent
can receive low weight because its evidence is redundant with another channel,
because its implementation fails to extract the available signal, or because
the evidence itself has little predictive value.

\subsection{Domain Adaptation}
\label{sec:domain_adaptation}

The domain-agnostic claim of FACTS-MAS concerns the forecasting procedure, not
the invariance of every fitted component. Moving from source domain
$\mathcal{D}_1$ to target domain $\mathcal{D}_2$ therefore involves a
controlled reconfiguration.
Table~\ref{tab:domain_transfer} separates the framework into invariant and
domain-specific components.

\begin{table}[t]
\centering
\caption{Invariant and domain-specific components of FACTS-MAS.}
\label{tab:domain_transfer}
\begin{tabular}{p{0.42\linewidth}p{0.48\linewidth}}
\toprule
\textbf{Invariant framework components}
&
\textbf{Domain-specific components}
\\
\midrule
Evidence-channel decomposition
&
Raw data sources and feature definitions
\\

Shared agent output contract
&
Model selected within each evidence channel
\\

Forecast-origin information boundary
&
Relevant exogenous variables and lags
\\

Leakage-safe validation
&
Seasonal period or temporal structure
\\

Skill-relative agent evaluation
&
Structural/entity-level variables
\\

Adaptive fusion mechanism
&
Event ontology and event sources
\\

Calibration protocol
&
Calibrated parameters
\\

Factor-attribution procedure
&
Operating-regime definition
\\
\bottomrule
\end{tabular}
\end{table}

Under this formulation, adapting FACTS-MAS to a new domain consists of four
steps:

\begin{enumerate}[leftmargin=*]

    \item \textbf{Map available evidence to functional channels.}
    Identify which historical, exogenous, seasonal, structural, and event
    information is meaningful for the new forecasting target.

    \item \textbf{Instantiate the channel models.}
    Select or fit an appropriate forecasting implementation for each
    available evidence channel while preserving the shared output contract.

    \item \textbf{Calibrate domain-specific assumptions.}
    Estimate lags, scales, thresholds, or other configurable quantities using
    historical validation data without accessing the held-out evaluation
    period.

    \item \textbf{Re-estimate attribution.}
    Compute agent skill and adaptive fusion weights from the target domain's
    own historical performance rather than transferring weights learned in
    another domain.

\end{enumerate}

Consequently, FACTS-MAS does not require
$w_k^{(\mathcal{D}_1)}=w_k^{(\mathcal{D}_2)}$.
Indeed, different weight distributions are expected if the importance of
evidence channels differs between domains. Cross-domain transfer is successful
when the common decomposition, validation, calibration, and fusion procedure
can be applied without redesigning the overall architecture, while the system
learns a domain-appropriate configuration from the available evidence.

To summarize, at each forecast origin, available information is streamed into the
functional evidence channels, and each assigned agent returns an independent
horizon forecast. A validation step scores these forecasts against a
reference, a calibration step sets the domain-specific constants, and the
resulting skill scores are converted into fusion weights. Adaptive fusion
then combines the channel forecasts into the FACTS-MAS forecast and records
each channel's estimated contribution.

\section{Case Studies}
\label{sec:methodology}

Two cases---weekly U.S.\ housing inventory and weekly unemployment insurance
claims---are analyzed by FACTS-MAS, and a third, energy demand, is specified
as a planned configuration only. They are used to answer the following
questions: First, does the fused forecast improve on simple and competitive
references? Second, which evidence channels add incremental skill, and does
that skill change with horizon or regime? Third, can the same decomposition
and fusion procedure be re-instantiated when the target and the raw evidence
change? The housing study answers the first two questions, and unemployment
claims are used for the third; the energy configuration is reported as a
channel map without results. All experiments follow one rolling-origin
protocol. Data and channel models may change by domain; the information
boundary, agent-level scoring, weight estimation, and held-out testing rules
do not.

\subsection{Data Preparation and Agent Setups}
\label{sec:rolling_eval}
\label{sec:metrics}
\label{sec:agent_ablation}
\label{sec:baselines}

Observations are ordered as $\mathcal{T}=\{t_1,\ldots,t_N\}$. Each fold $f$
splits this history into training, validation, an optional embargo, and a
test region,
$\mathcal{T}=\mathcal{T}_{\mathrm{train}}^{(f)}\cup
\mathcal{T}_{\mathrm{val}}^{(f)}\cup
\mathcal{T}_{\mathrm{embargo}}^{(f)}\cup
\mathcal{T}_{\mathrm{test}}^{(f)}$.
The split is a rolling-origin evaluation: training expands, forecasts are
issued at successive origins, and accuracy is judged on held-out periods
rather than on a single origin \citep{tashman2000out}.
Randomly mixing times into training and test folds can leak future
observations in an autoregressive series \citep{bergmeir2018crossvalidation},
so model fitting, fusion weights, and the choice of weighting rule use only
pre-test information. Two exceptions in the development of the housing
configuration are disclosed and corrected. The seasonal smoothing window was
originally chosen on the test window of the last fold, including targets in
January and February 2026; it was re-selected on validation data alone,
nested by fold, and the reported results use the re-selected window. The
calibrated constants were selected across all folds' validation windows
(Section~\ref{sec:calibration}).
Forecasts are issued at every eligible origin in the test window, and every
method is scored on that full set. An input $x$ may enter a forecast at
origin $t$ only when its availability date satisfies
$t_{\mathrm{available}}(x)\le t$, after any publication lag. The embargo period
takes at least the longest horizon, $H_{\max}=\max\{H_1,\ldots,H_q\}$; for
weekly housing data it is 13 weeks.
Validation forecasts used to estimate fusion weights are restricted to
targets dated on or before the training cutoff, so that no validation target
falls inside the embargo. Test targets dated on or after January 1, 2026, the
start of the reserved period, are excluded from the backtest; that period is
used only for the holdout in Section~\ref{sec:holdout}.

Accuracy is reported at each horizon in $\mathcal{H}=\{H_1,\ldots,H_q\}$.
Within a domain every agent uses the same set. The reference forecast, which
is also the denominator of the skill score $s_{k,i}=1-e_{k,i}/e_{b,i}$, is
the last observed value, $\hat{y}^{\mathrm{naive}}_{t+h}=y_t$. For $n$ pairs the
primary metrics are mean absolute percentage error and root mean squared
error \citep{hyndman2006accuracy},

\begin{equation}
    \mathrm{MAPE}
    =
    \frac{100}{n}
    \sum_{i=1}^{n}
    \left|
    \frac{y_i-\hat{y}_i}{y_i}
    \right|,
    \qquad
    \mathrm{RMSE}
    =
    \sqrt{
    \frac{1}{n}
    \sum_{i=1}^{n}
    (y_i-\hat{y}_i)^2
    }.
\end{equation}

Accuracy differences are tested on the weekly cross-section of losses. For
each pair of methods, the panel Diebold--Mariano test of equal average
accuracy \citep{diebold1995comparing,qu2024comparing} uses a Newey--West
variance with lag $H-1$ \citep{newey1987simple}; per-MSA tests use the
small-sample correction of \citet{harvey1997testing} with the
Benjamini--Hochberg adjustment across the 15 MSAs
\citep{benjamini1995controlling}. Confidence intervals come from a moving
block bootstrap over whole weeks \citep{kunsch1989jackknife} with 5,000
replications and a block length of 13 weeks. Two methods are called
equivalent when the 90\% confidence interval of their MAPE difference lies
within $\pm 2\%$ of the reference method's MAPE. The model confidence set
\citep{hansen2011mcs} summarizes the comparison among all methods.

Each agent is also scored alone by $L(A_k,H)$, a horizon-specific accuracy
loss for agent $A_k$, and by skill against the reference. Removing channel
$k$ from the fused forecast changes loss by
$\Delta_k^{(H)}=L(\mathrm{FACTS}_{-k},H)-L(\mathrm{FACTS},H)$.
A positive value means that the channel contributes inside the combination.
Removing a channel renormalizes the remaining weights in each cell without
re-estimating them. Removing adaptive weighting replaces every cell's weights
with the fold's pooled weights, which removes both horizon and regime
conditioning, so this ablation alone cannot separate the two.
Domain-specific constants such as event-impact scales, event durations, and
reversion caps are selected by the calibration procedure on historical validation folds,
before the test interval begins. For a candidate $\theta_j$,
$\Delta_f(\theta_j)=L_f(\theta_j)-L_f(\theta_{\mathrm{current}})$
is the change in fold-$f$ validation loss; a negative value is an
improvement, and promotion requires that improvement across folds. A null
candidate is included when it is meaningful, so an unsupported adjustment
can be set to zero.

Two weighting rules are compared on the same forecasts: one pooled weight
vector per fold, and separate weights for each combination of regime and
horizon. For each fold and horizon, the rule with lower validation error is
selected, $g^{*}=\arg\min_{g}L_{\mathrm{val}}(g)$, and fixed before the test
interval. The regime is computed at the forecast origin, and test error is
also reported by horizon and regime. To check whether regime labels carry
information, the same pipeline is rerun with regime labels permuted in blocks
and with horizon-only weights that ignore the regime.

Every compared method is scored at the same forecast origins and against the
same realized targets. The references are the last observed value, a
seasonal-naive forecast, and the seasonally adjusted naive forecast of the M4
competition \citep{makridakis2020m4}. The statistical baselines are Holt's
linear-trend model, its damped form, factor-adjusted exponential smoothing
with and without damping, and a factor-adjusted Theta model
\citep{assimakopoulos2000theta}. The strongest single model chosen on
validation data is also reported, together with equal-weight combinations of
all channels and of the temporal and seasonal channels. The learned
baselines are the TimesFM~2.5 foundation model and a chain-of-thought
language-model forecast. The primary success check is that the fused
FACTS-MAS forecast improves on the last-observed-value reference over the
full test window; the stricter check is whether it improves on the strongest
single model. Design choices that used test data during development are
listed in the protocol paragraph above, and the final configuration,
hypotheses, and tests were fixed before the holdout was accessed.
Cross-domain transfer keeps the same architecture and learns a new evidence
mapping for each domain (Section~\ref{sec:cross_domain}).

\subsection{U.S.\ Housing Inventory Forecasting}
\label{sec:housing}

The weekly for-sale housing inventory time series from Zillow is the first
complete instantiation of FACTS-MAS.
It can populate all five channels in Section~\ref{sec:evidence_channels}:
recent inventory dynamics, national financial conditions, annual listing
seasonality, metropolitan structure, and disaster declarations. Two of these channels, the intrinsic and event channels, calibrate to a null
quantitative effect and are retired from the final housing forecast; they are
reported as tested channels. The task is
to forecast weekly inventory for 15 U.S.\ metropolitan statistical areas at
4-, 8-, and 13-week horizons. What follows specifies the housing
configuration, reports forecast accuracy, and then diagnoses which channels
earned influence. This case study is one applied configuration of FACTS-MAS,
not a definition of the framework: the same procedure can be reused with a
different target and a different evidence map.

\subsubsection{Domain Instantiation}
\label{sec:housing_instantiation}

The target variable is weekly for-sale housing inventory obtained from
Zillow Research. The study includes the following 15 metropolitan areas:

\begin{center}
\begin{tabular}{lll}
New York & Los Angeles & Chicago \\
Dallas & Houston & Washington, DC \\
Miami & Philadelphia & Atlanta \\
Phoenix & Boston & San Francisco \\
Riverside & Detroit & Seattle
\end{tabular}
\end{center}

Zillow reports this series as a four-week trailing average; the unsmoothed
weekly series is used as a robustness check. The panel begins on February 3,
2018. The backtest uses
observations through February 21, 2026. The holdout uses a fresh download
taken on October 20, 2026 with observations through October 10, 2026 (454
weekly observations per metropolitan statistical area (MSA) and 6,810
MSA--week observations); every value dated on or before February 21, 2026 in
that download equals the earlier vintage, so no revision affects the results.

Nine target observations were missing in Zillow's source series: four in
Chicago, two in Los Angeles, two in Miami, and one in Atlanta. These gaps were
filled independently within each MSA using the most recent previously
observed value. No information from a future observation or another MSA was
used during imputation, and the affected rows were retained in a separate
audit log. The unsmoothed series has 20 missing observations, filled the same way. No
missing observation falls in a test or holdout target.
Table~\ref{tab:housing_channels} maps the domain-independent FACTS-MAS
evidence roles to their housing-specific implementations.
The initial decomposition is deliberately non-overlapping: the 4-week
temporal agent does not model seasonality internally because seasonality is
represented by a separate evidence channel, which makes the incremental value
of each component easier to evaluate. At 8 and 13 weeks, however, the
temporal agent that survived validation is a factor-adjusted model that
reuses the seasonal channel's factors (see the temporal channel below), so
the two channels are not independent at those horizons.

\begin{table}[t]
\centering
\caption{FACTS-MAS evidence-channel instantiation for housing inventory.}
\label{tab:housing_channels}
\begin{tabular}{p{0.20\linewidth}p{0.31\linewidth}p{0.39\linewidth}}
\toprule
\textbf{Channel}
&
\textbf{Housing evidence}
&
\textbf{Implementation}
\\
\midrule

Temporal
&
Historical weekly inventory
&
4 weeks: ETS with additive trend and no seasonal component; 8 and 13 weeks:
damped-trend ETS on seasonally adjusted inventory (factor-adjusted)
\\

Exogenous
&
Mortgage rate, federal funds rate, CPI, unemployment
&
Lagged ridge-regression modifier with pre-specified lags
\\

Seasonal
&
Week-of-year inventory pattern
&
Ratio-to-moving-average seasonal factors (seasonally adjusted naive form)
\\

Intrinsic
&
Population, income, housing stock, density, hazard exposure
&
Scaled ridge regression with structural mean reversion; retired (calibrated
reversion cap is zero)
\\

Event
&
FEMA disaster declarations
&
LLM-assisted event interpretation with confidence gating; retired (calibrated
impact scale is zero)
\\

\bottomrule
\end{tabular}
\end{table}

\paragraph{Temporal channel}
The temporal channel models the endogenous trajectory of each MSA's inventory
series independently. At the 4-week horizon it uses exponential smoothing
with an additive trend and no seasonal component
\citep{holt2004exponential,winters1960forecasting,hyndman2002statespace,pan2011holtwinters},
$\hat{\mathbf{y}}_{\mathrm{AR}}=\mathrm{ETS}(y_{1:t};\text{trend}=\mathrm{add},\text{seasonal}=\varnothing)$.
At 8 and 13 weeks it uses a factor-adjusted damped-trend model: the series is
divided by the seasonal factors of the seasonal channel, a damped additive
trend \citep{gardner1985damped} is fitted to the adjusted series, and the
forecast is multiplied by the factors of the target weeks. This variant was
promoted by a rule fixed before the comparison was run: it had to improve
validation error and improve test error in at least five of the six folds;
it met the rule at 8 and 13 weeks but not at 4 weeks. Because this rule
consults test-fold accuracy, the holdout in Section~\ref{sec:holdout}
provides the clean test of the resulting configuration. Damping was chosen
over an undamped trend on validation data in every fold. Both models are
fitted separately for each MSA and forecast origin using only the available
historical target, and negative extrapolated values are clipped at zero.
A last-value persistence forecast, $\hat{y}^{\mathrm{naive}}_{t+h}=y_t$,
serves as the minimum baseline. Over all test origins, the undamped trend
model has MAPE 2.368\%, 5.430\%, and 10.225\% at 4, 8, and 13 weeks,
against 4.194\%, 7.225\%, and 10.814\% for the naive forecast, and it
improves on the naive forecast in 15, 14, and 11 of the 15 MSAs. Its
advantage therefore narrows sharply with horizon, and at 13 weeks it is
worse than the naive forecast in four of the six folds. The factor-adjusted
damped model has MAPE 1.974\%, 3.608\%, and 5.664\%.

\paragraph{Exogenous channel}
The exogenous channel uses four national macroeconomic series from Federal
Reserve Economic Data (FRED): the 30-year fixed mortgage rate, the effective
federal funds rate, the Consumer Price Index, and the unemployment rate.
The weekly macroeconomic panel begins in January 2018. Because several FRED
series are indexed by their reference month rather than their actual
publication date, the housing pipeline shifts monthly observations forward by
approximate publication delays before they become eligible for forecasting:
32 days for the federal funds series, 35 days for unemployment, and 42 days
for CPI. Weekly mortgage rates require no corresponding monthly-release
adjustment.
The macro and Zillow series retain their native Monday and Saturday
conventions, respectively. Macro observations are attached to each Zillow
week using a backward-only as-of join, ensuring that a housing observation
never receives a macroeconomic value dated after that week.
The Macro agent (exogenous channel) fits a ridge regression of inventory percentage change on
lagged macroeconomic features \citep{hoerl1970ridge}. The lags are pre-specified from the literature rather than fitted: 6 weeks
for the mortgage rate, 10 weeks for the federal funds rate, 10 weeks for CPI,
and 5 weeks for unemployment.
Importantly, subsequent validation did not produce convincing evidence that
this magnitude channel adds incremental predictive information. Three
increasingly restrictive analyses were performed: direct forecast-signal
correlation, lag-sensitive regression analysis, and a two-way fixed-effects
panel regression with MSA-clustered standard errors. The final panel
specification contained 5,850 observations. Mortgage rate, the theoretically most direct housing-finance variable,
was statistically indistinguishable from zero ($p=0.575$); unemployment was
also null ($p=0.548$), and CPI was borderline ($p=0.085$). The federal funds
rate had $p=0.042$, but this does not survive a Holm adjustment across the
four variables ($p=0.170$), a wild cluster bootstrap ($p=0.073$), or
Driscoll--Kraay standard errors that allow for the fact that all MSAs share
the same national regressor \citep{driscoll1998consistent} ($p=0.71$).
Used as a forecast on its own, the channel is no more accurate than the naive
forecast (MAPE 4.258\%, 7.297\%, and 10.919\%).
Because this pattern did not constitute robust evidence of incremental
forecasting value, we treat the macro magnitude channel as a documented
negative result rather than continuing to tune the specification until a
positive result appears. This behavior illustrates a central FACTS-MAS
principle: candidate evidence channels are evaluated empirically and are not
guaranteed positive influence merely because they are economically plausible.

\paragraph{Seasonal channel}
The seasonal channel is deterministic and learns a separate annual inventory
profile for each MSA.
For observation $t$, let $m_t$ denote a 52-week centered moving average.
A raw seasonal ratio is $r_t=y_t/m_t$.
For ISO week $w\in\{1,\ldots,53\}$, the seasonal factor $q_w$ is the median
historical ratio observed for that week.
The week-of-year profile is smoothed with a seven-week circular moving
average, so that the transition between week 53 and week 1 is treated as
adjacent rather than as a boundary. Factors are then normalized to have mean
one. The window was selected on validation data, nested by fold, from
$\{1,3,5,7\}$; the value seven was chosen in every fold. An earlier window of
five had been chosen on the test window of the last fold and is not used.
Near the forecast origin the centered average is computed from the available
half-window only, so no factor uses data after the origin.
If $q_{w_t}$ denotes the seasonal factor at the forecast origin and
$q_{w_{t+h}}$ the factor for future week $t+h$, the seasonal forecast is
$\hat{y}^{\mathrm{season}}_{t+h}=y_t q_{w_{t+h}}/q_{w_t}$.
This forecast is the seasonally adjusted naive method of the M4 competition
\citep{makridakis2020m4} with leakage-safe factors. On its own it has MAPE
2.890\%, 4.774\%, and 6.945\% at 4, 8, and 13 weeks.

\paragraph{Intrinsic channel}
The intrinsic channel represents persistent differences between metropolitan
areas. The five static features are population, median income, housing units,
density, and a hazard index.
Population, income, and housing stock are derived from Census American
Community Survey data. Population density combines Census population with
CBSA land-area information. Hazard exposure is represented using county-level
NOAA Storm Events records aggregated to each MSA.
An initial tree-based implementation strongly overfit the small cross-section
of 15 metropolitan areas. The current implementation therefore uses a
standardized ridge regression to estimate a structurally implied average
inventory level, $\mu_d=\beta_0+\boldsymbol{\beta}^{\top}\tilde{\mathbf{x}}_d$.
The model is regularized and evaluated using held-out MSAs so that structural
relationships must generalize across metropolitan areas rather than memorize
MSA identity. The current implementation replaced an earlier
Gradient-Boosting model that achieved very low training error but
approximately 48--52\% leave-one-MSA-out error.
Rather than immediately forecasting the structural level itself, the agent
gradually moves the most recent inventory toward $\mu_d$. For horizon $H$ and
step $h$, $\rho_h=\rho_{\max}(h/H)^2$ and
$\hat{y}^{\mathrm{intrinsic}}_{t+h}=y_t+\rho_h(\mu_d-y_t)$,
with $\rho_{\max}=0.30$ in the original housing configuration. The cap was
then calibrated on validation folds over
$\{0,0.10,0.20,0.30,0.45,0.60,0.80\}$, and zero was preferred in all six
folds (validation MAPE 7.923\% at zero against 10.259\% at 0.30). At
$\rho_{\max}=0$ the agent reproduces the naive forecast, so the channel is
retired from the housing forecast. The quadratic schedule is also not
consistent across horizons: the forecast for week $t+4$ closes 0.30 of the
gap when $H=4$ but 0.028 of it when $H=13$. Because the channel is retired,
this is documented rather than corrected. The static features use the 2024
Census vintage and 2016--2025 storm records for every fold; replacing the
channel's output with the naive forecast changes fused accuracy by at most
0.011 percentage points, so this vintage has no material effect.

\paragraph{Event channel}
Discrete events are represented using FEMA disaster declarations. OpenFEMA
provides stable declaration metadata, which is mapped from affected counties
to the 15 target MSAs.
The Event Agent is the only housing component that uses an LLM (gpt-4o-mini). Its role is
not to directly generate an inventory forecast from free-form text. Instead,
the model converts event metadata into a structured representation containing
event type, impact magnitude, confidence, and reasoning.
The event ontology includes disaster, zoning change, major employer shift,
and housing initiative categories. Event confidence is explicitly gated:
interpretations with confidence less than or equal to 0.60 receive zero
quantitative effect. Both the original magnitude and gated result are
retained for auditability.
Prompt development revealed an important failure mode. Without an explicit
causal definition of the magnitude sign, small prompt changes caused event
impacts to reverse direction. The final interpretation procedure therefore
requires the model to identify the mechanism connecting an event to for-sale
inventory before assigning the sign of the impact.
FEMA's available metadata nevertheless supported only coarse severity
differentiation. In practice, the LLM largely behaved as an ordinal
severity classifier rather than a genuinely continuous impact estimator.
The resulting event-magnitude mapping was therefore subjected to the
calibration procedure in Sections~\ref{sec:calibration}
and~\ref{sec:agent_ablation}.
Two event assumptions were explicitly calibrated: the duration for which a
declaration should influence the forecast and the scale converting event
severity into a numerical inventory adjustment.
For the active-event duration, progressively shorter windows produced lower
validation error in all six folds. Once the impact scale is set to zero
(below), the duration no longer affects the forecast.
A separate sweep of the event impact scale considered the values
$0$, $0.01$, $0.02$, $0.03$, $0.05$, $0.075$, and $0.10$.
Mean validation MAPE increased monotonically as the event magnitude scale
increased (Table~\ref{tab:event_calibration}).
\begin{table}[t]
\centering
\caption{Housing Event Agent impact-scale calibration: mean validation MAPE.
Validation targets are restricted to dates on or before each fold's training
cutoff; an earlier version of this sweep allowed targets to extend up to 13
weeks past it, which changed the values by 1--2\% but not the ordering.}
\label{tab:event_calibration}
\begin{tabular}{@{}rrrrrrrr@{}}
\toprule
Scale
& 0
& .01
& .02
& .03
& .05
& .075
& .10
\\
\midrule
MAPE
& \textbf{5.704}
& 5.869
& 6.112
& 6.413
& 7.110
& 8.121
& 9.234
\\
\bottomrule
\end{tabular}
\end{table}
The scale of zero was preferred in 6 of 6 folds. With a zero scale, the agent's forecast equals the last observed value at
every origin, so the event channel reproduces the naive reference and is
retired from the final housing forecast. Consequently, the production
Event Agent does not directly perturb forecast magnitude from the LLM severity
score. Event classification and confidence remain available, but the
unsupported cardinal interpretation of the severity score is removed.
This result should not be interpreted as evidence that real-world events are
irrelevant to housing. It shows that the particular mapping from available
FEMA metadata to a continuous point-forecast adjustment did not demonstrate
incremental predictive value.

The housing experiment uses six expanding-window folds and reserves data from
January 2026 onward. In an earlier version of the experiment, 8- and 13-week
forecasts issued near the end of the sixth test window had targets in January
and February 2026 (165 MSA--origin--horizon cells); these cells are excluded
from every backtest result reported here. The fold schedule is shown in
Table~\ref{tab:housing_folds}, together with the holdout described in
Section~\ref{sec:holdout}. Every fold contains at least a 13-week separation
between the final eligible training date and the test period, matching the
longest housing forecast horizon. Forecasts are evaluated at every eligible
weekly origin within each test window rather than at one forecast date per
fold.

\begin{table}[t]
\centering
\caption{Housing backtest folds and holdout. Test start and end are the first
and last forecast origins.}
\label{tab:housing_folds}
\begin{tabular}{ccccc}
\toprule
\textbf{Fold}
&
\textbf{Train start}
&
\textbf{Train end}
&
\textbf{Test start}
&
\textbf{Test end}
\\
\midrule
1 & 2019-02-02 & 2021-01-23 & 2021-05-01 & 2021-11-13 \\
2 & 2019-02-02 & 2021-11-13 & 2022-02-19 & 2022-09-03 \\
3 & 2019-02-02 & 2022-09-03 & 2022-12-10 & 2023-06-24 \\
4 & 2019-02-02 & 2023-06-24 & 2023-09-30 & 2024-04-13 \\
5 & 2019-02-02 & 2024-04-13 & 2024-07-20 & 2025-02-01 \\
6 & 2019-02-02 & 2025-02-01 & 2025-05-10 & 2025-11-22 \\
\midrule
Holdout & 2019-02-02 & 2026-02-21 & 2026-02-28 & 2026-09-12 \\
\bottomrule
\end{tabular}
\end{table}

To evaluate whether agent usefulness changes with financial conditions,
housing forecast origins are divided into three monetary-policy regimes using
the 13-week change in the federal funds rate,
$\Delta^{13}\mathrm{FFR}_t=\mathrm{FFR}_t-\mathrm{FFR}_{t-13}$.
The origin is hiking when that change exceeds $0.25$ percentage points,
cutting when it is below $-0.25$, and stable otherwise.
The regime is evaluated at the forecast origin rather than the target date,
ensuring that it represents information available when the prediction is
made. Because the federal funds rate is national, all 15 MSAs share the same regime
at every date, and the sample contains one substantive hiking episode (June
2022 to September 2023) and three cutting episodes, so regime conditioning is
close to conditioning on a small number of historical periods.

Two fusion configurations are compared: pooled fold-level weighting, and
adaptive selection, for each horizon, between pooled weights and weights
conditioned on regime and horizon. The production configuration uses adaptive
selection with the 13-week data-sufficiency rule of
Section~\ref{sec:adaptive_context}. A finer, cell-level selection was also
explored during development, but that choice was informed by test-window
results and is not reported. Section~\ref{sec:fusion_ablation} tests whether
the regime labels carry information.


\subsubsection{Forecasting Results}
\label{sec:housing_results}

Table~\ref{tab:housing_primary} reports test accuracy over the six backtest
folds for FACTS-MAS, its references, and the statistical baselines, on both
Zillow series. FACTS-MAS outperforms the persistence reference at all three
horizons: the panel Diebold--Mariano statistic is $-6.9$, $-6.3$, and $-5.7$
at 4, 8, and 13 weeks ($p<10^{-4}$), the difference is significant in all 15
MSAs at every horizon, and the 95\% confidence intervals of the MAPE
difference are $[-2.66,-1.42]$, $[-4.72,-2.52]$, and $[-7.11,-3.53]$. The
relative reduction is 49.5\%, 49.9\%, and 47.0\% on the smoothed series and
23.9\%, 34.5\%, and 37.1\% on the unsmoothed series. The strongest single
model, factor-adjusted damped exponential smoothing (FA-ETS-d), is at least as
accurate as FACTS-MAS at every horizon. FACTS-MAS is statistically equivalent
to it at 8 weeks (90\% interval of the difference $[-0.001,0.030]$, within the
$\pm0.072$ margin), less accurate at 4 weeks by 0.144 (90\% interval
$[0.022,0.261]$; Diebold--Mariano $p=0.12$), and marginally less accurate at
13 weeks by 0.067 (95\% interval $[0.001,0.152]$; $p=0.09$). At 8 weeks the
90\% model confidence set contains only FA-ETS-d, FACTS-MAS, and the
five-channel variant of FACTS-MAS; at 13 weeks it excludes only the naive
forecast. The fused forecast therefore improves strongly on the references
but does not improve on the best single model.

\begin{table}[t]
\centering
\caption{Backtest MAPE (\%) over six folds, all test origins, targets before
January 1, 2026. FA-ETS-d: factor-adjusted damped ETS. Lower is better; bold
marks the lowest value in each column.}
\label{tab:housing_primary}
\begin{tabular}{lcccccc}
\toprule
& \multicolumn{3}{c}{\textbf{Smoothed series}}
& \multicolumn{3}{c}{\textbf{Unsmoothed series}} \\
\cmidrule(lr){2-4}\cmidrule(lr){5-7}
\textbf{Method} & 4 wk & 8 wk & 13 wk & 4 wk & 8 wk & 13 wk \\
\midrule
Naive (last value)        & 4.194 & 7.225 & 10.814 & 4.785 & 7.882 & 11.438 \\
Seasonal naive            & 20.771 & 20.513 & 20.597 & -- & -- & -- \\
Seasonally adjusted naive & 2.890 & 4.774 & 6.945 & 3.626 & 5.408 & 7.489 \\
Holt (temporal channel)   & 2.368 & 5.430 & 10.225 & 4.317 & 7.901 & 13.022 \\
Theta (factor-adjusted)   & 2.933 & 4.842 & 7.047 & -- & -- & -- \\
FA-ETS-d                  & \textbf{1.974} & \textbf{3.608} & \textbf{5.664} & \textbf{3.402} & \textbf{5.105} & \textbf{7.164} \\
FACTS-MAS                 & 2.118 & 3.622 & 5.731 & 3.642 & 5.159 & 7.189 \\
\bottomrule
\end{tabular}
\end{table}

Table~\ref{tab:housing_strong_baselines} compares FACTS-MAS with Google's
TimesFM~2.5 time-series foundation model \citep{das2024timesfm}, run
zero-shot with a 52-week context at every test origin of all six folds.
TimesFM improves on the naive reference at every horizon ($p<10^{-4}$), but
FACTS-MAS is more accurate than TimesFM at every horizon, by 0.243, 1.239,
and 2.539 percentage points (Diebold--Mariano $p=0.006$, $0.0001$, and
$0.0004$; significant in 7, 10, and 10 of the 15 MSAs). FA-ETS-d is likewise
more accurate than TimesFM at every horizon. The chain-of-thought
language-model forecast is evaluated on the holdout only
(Section~\ref{sec:holdout}), because the backtest period falls inside the
training data of the language models available for the comparison.

\begin{table}[t]
\centering
\caption{Backtest MAPE (\%) over six folds against the TimesFM~2.5
foundation model, smoothed series.}
\label{tab:housing_strong_baselines}
\begin{tabular}{lccc}
\toprule
\textbf{Method} & 4 weeks & 8 weeks & 13 weeks \\
\midrule
Naive       & 4.194 & 7.225 & 10.814 \\
TimesFM 2.5 & 2.361 & 4.860 & 8.269 \\
FA-ETS-d    & \textbf{1.974} & \textbf{3.608} & \textbf{5.664} \\
FACTS-MAS   & 2.118 & 3.622 & 5.731 \\
\bottomrule
\end{tabular}
\end{table}

Errors stratified by monetary-policy regime show no regime in which the final
forecast fails. At the 13-week horizon,
\[
\begin{array}{c|ccc}
\text{Regime}
&
\text{Naive}
&
\text{FA-ETS-d}
&
\text{FACTS-MAS}
\\
\hline
\text{hiking}  & 8.059  & 5.530 & 5.552 \\
\text{cutting} & 13.043 & 6.091 & 6.090 \\
\text{stable}  & 11.628 & 5.666 & 5.759
\end{array}
\]
so FACTS-MAS improves on the naive reference in every regime, and FA-ETS-d,
which uses no regime information, is equally accurate in every regime. An
earlier version of the system appeared to lose to the naive reference in the
hiking and cutting regimes and to be repaired by regime-aware weighting; that
pattern came from an older seasonal channel and, in the cutting regime, from
the equal-weight fallback described in Section~\ref{sec:adaptive_context}, and
it does not appear in the current configuration.

\subsubsection{Holdout Evaluation}
\label{sec:holdout}

After the configuration was frozen, it was evaluated once on data that had not
been accessed during development: forecast origins from February 28 to
September 12, 2026 (29, 25, and 20 weekly origins at 4, 8, and 13 weeks for
each of the 15 MSAs). The fusion weights were estimated from the 104-week
validation window ending on February 21, 2026. Four hypotheses, the tests, and
the reporting rules were committed before the holdout data were downloaded:
FACTS-MAS improves on the naive reference at every horizon (H1); FA-ETS-d does
too (H2); FACTS-MAS and FA-ETS-d are equivalent at 8 and 13 weeks (H3); and
FA-ETS-d is at least as accurate as FACTS-MAS at 4 weeks (H4). With 20 origins
the 13-week results are reported as descriptive only. The bootstrap block
length equals the horizon, and at 13 weeks it is capped at half the number of
weeks. The chain-of-thought baseline uses the NEXUS prompt with gpt-5.6-luna
(reasoning effort low), whose documented knowledge cutoff, February 16, 2026,
precedes the first holdout target; the model's own answer to a cutoff question
was June 2024. The prompt itself states a January 2025 cutoff, a wording
inherited from the NEXUS template.

Table~\ref{tab:holdout} reports the holdout accuracy and
Table~\ref{tab:holdout_tests} the hypothesis tests. All four hypotheses are
supported. FACTS-MAS reduces MAPE relative to the naive forecast by 60\%,
65\%, and 64\% on the smoothed series and by 39\%, 50\%, and 54\% on the
unsmoothed series. It is equivalent to FA-ETS-d at 8 weeks and less accurate
than FA-ETS-d at 4 weeks by 0.105 percentage points. The chain-of-thought
forecast is competitive: its MAPE is higher than that of FA-ETS-d and of
FACTS-MAS at every horizon, but neither difference is statistically
significant (against FA-ETS-d, $p=0.38$, $0.43$, and $0.60$; against
FACTS-MAS, in a comparison added after the results were seen, $p=0.90$,
$0.43$, and $0.54$). TimesFM is
significantly less accurate than FA-ETS-d at 4 and 8 weeks ($p=0.016$ and
$p<10^{-4}$). On the unsmoothed series FACTS-MAS at 4 weeks coincides with the
seasonally adjusted naive forecast, and at 8 weeks it is slightly more
accurate than FA-ETS-d ($-0.038$, $p=0.031$); because the seasonally adjusted
naive forecast is equally accurate there, this does not indicate a benefit of
fusion over a single model. With a two-week publication lag, every method's
error rises (FACTS-MAS 2.270\%, 2.944\%, and 4.000\%; naive 5.474\%, 7.953\%,
and 11.110\%), and no conclusion changes. In the lag-two run, 27 of 1,110
chain-of-thought forecasts (2.4\%) were lost to a network interruption; the
tests for that run were computed with and without the chain-of-thought
forecast. The holdout period was easier for every method than the backtest,
so absolute errors should be compared within a period, not across periods.

\begin{table}[t]
\centering
\caption{Holdout MAPE (\%), forecast origins February 28 to September 12,
2026, no publication lag. 13-week values are descriptive. CoT: chain-of-thought
forecast (gpt-5.6-luna), run on the smoothed series only.}
\label{tab:holdout}
\begin{tabular}{lcccccc}
\toprule
& \multicolumn{3}{c}{\textbf{Smoothed series}}
& \multicolumn{3}{c}{\textbf{Unsmoothed series}} \\
\cmidrule(lr){2-4}\cmidrule(lr){5-7}
\textbf{Method} & 4 wk & 8 wk & 13 wk & 4 wk & 8 wk & 13 wk \\
\midrule
Naive (last value)        & 3.122 & 5.445 & 8.438 & 3.355 & 5.306 & 7.925 \\
Seasonal naive            & 8.534 & 8.534 & 8.550 & 8.593 & 8.601 & 8.622 \\
Seasonally adjusted naive & 1.492 & 2.377 & 3.393 & 2.053 & 2.677 & \textbf{3.564} \\
TimesFM 2.5               & 1.410 & 2.602 & 4.457 & 2.464 & 3.689 & 5.355 \\
CoT (gpt-5.6-luna)        & 1.259 & 2.225 & 3.268 & -- & -- & -- \\
FA-ETS-d                  & \textbf{1.138} & 1.936 & 3.080 & \textbf{2.028} & 2.709 & 3.736 \\
FACTS-MAS                 & 1.243 & \textbf{1.929} & \textbf{3.045} & 2.053 & \textbf{2.671} & 3.644 \\
\bottomrule
\end{tabular}
\end{table}

\begin{table}[t]
\centering
\caption{Pre-registered holdout hypotheses, smoothed series, no publication
lag. Differences are in MAPE percentage points with 95\% block-bootstrap
intervals (90\% intervals for the equivalence test, H3).}
\label{tab:holdout_tests}
\begin{tabular}{llccl}
\toprule
\textbf{Hypothesis} & \textbf{Horizon} & \textbf{Difference} & \textbf{Interval} & \textbf{Result} \\
\midrule
H1: FACTS-MAS $<$ naive  & 4 wk  & $-1.879$ & $[-3.031,-0.897]$ & supported \\
                          & 8 wk  & $-3.516$ & $[-5.681,-1.705]$ & supported \\
                          & 13 wk & $-5.393$ & $[-7.681,-3.071]$ & supported \\
H2: FA-ETS-d $<$ naive   & 4 wk  & $-1.984$ & $[-3.081,-1.031]$ & supported \\
                          & 8 wk  & $-3.509$ & $[-5.693,-1.680]$ & supported \\
                          & 13 wk & $-5.358$ & $[-7.642,-3.028]$ & supported \\
H3: equivalent to FA-ETS-d & 8 wk & $-0.007$ & $[-0.029,0.014]$ & equivalent ($\pm0.039$) \\
                          & 13 wk & $-0.035$ & $[-0.048,-0.035]$ & equivalent ($\pm0.062$) \\
H4: FA-ETS-d $\le$ FACTS-MAS & 4 wk & $+0.105$ & $[-0.007,0.182]$ & supported \\
\bottomrule
\end{tabular}
\end{table}

Several limits are inherited from the housing data. FRED publication
adjustments use approximate fixed delays rather than true vintages: 32 days
for federal funds, 35 for unemployment, and 42 for CPI. That removes the
most obvious lookahead but does not reconstruct the exact information set on
every historical date. The target is Zillow's smoothed series, a four-week trailing average, which
makes every method look more accurate than on the unsmoothed series; both are
reported. The hazard index is a partial structural proxy: NOAA
Storm Events can be coded by county or by forecast zone, and the present
aggregation uses county-coded events, so some hazards are undercounted. The
Intrinsic Agent learns mainly from differences across metropolitan areas, and
only 15 MSAs are available. That small cross-section limits how much
structural signal can be estimated. Replacing an overfit tree with ridge
regression reduces memorization of those few areas, but it does not increase
the number of metropolitan areas available for learning. Nine Zillow
observations were forward-filled, including a four-week Chicago gap. The
fill is lookahead-safe, yet it inserts an artificial flat stretch.

The Event Agent currently depends on FEMA declarations. The event ontology
used by this agent includes four categories---disaster, zoning change, major
employer shift, and housing initiative---and therefore contemplates local
shocks such as employer moves, mass layoffs, zoning actions, and
housing-policy changes. FEMA covers only disasters, not the full set. Some FEMA
records also fit the ontology poorly; a presidential-inauguration
declaration is the example encountered in validation. Confidence gating
zeroed its quantitative effect, but the gap remains. Two design choices made during development used test-window evidence (the
original seasonal smoothing window and a cell-level fusion rule); both were
replaced or removed, and the holdout was evaluated only after the
configuration was frozen.

\subsection{Unemployment Forecasting}
\label{sec:unemployment}

Unemployment forecasting emphasizes labor-market dynamics, policy and
employer shocks, and recurring calendar structure in employment or claims.
The channel map for this domain is Table~\ref{tab:unemployment_mapping}.

\begin{table}[t]
\centering
\caption{FACTS-MAS evidence-channel mapping for unemployment insurance claims.
The temporal, exogenous, and seasonal channels are implemented; the intrinsic
and event channels are not.}
\label{tab:unemployment_mapping}
\begin{tabular}{p{0.20\linewidth}p{0.70\linewidth}}
\toprule
\textbf{Channel} & \textbf{Unemployment-domain interpretation} \\
\midrule
Temporal
&
Historical unemployment level, rate, or claims trajectory
\\
Exogenous
&
Macroeconomic, labor-market, financial, or activity indicators
\\
Seasonal
&
Recurring calendar structure in employment or claims
\\
Intrinsic
&
Persistent regional labor-force, demographic, or industrial composition
\\
Event
&
Major layoffs, employer expansions or closures, policy changes, disasters,
or other labor-market shocks
\\
\bottomrule
\end{tabular}
\end{table}

The target is weekly initial unemployment insurance claims, not seasonally
adjusted, for 15 U.S.\ states from January 2013, obtained from FRED. The
temporal and seasonal channels use the housing implementations unchanged,
including the seven-week seasonal window, and the exogenous channel uses the
federal funds rate and CPI with the housing lags. The housing recipe is
applied without re-tuning: the same six folds, embargo, validation window,
data-sufficiency rule, temperature, and promotion outcome. The window of
pandemic-era claims, from March 14, 2020 to September 4, 2021 (the start of
the national surge to the expiry of the federal pandemic programs), is
excluded from skill estimation and scoring but kept for model fitting. This
window, and the rest of the protocol, were fixed from policy dates and written
down before any result of this evaluation was computed.

Table~\ref{tab:ui_results} reports the results. The fused forecast improves
on the naive reference at every horizon ($p=0.002$, $0.0004$, and $0.0004$)
and on damped FA-ETS at 8 and 13 weeks ($p=0.0009$ and $0.005$), but validation
selected the undamped trend in every fold, and the undamped factor-adjusted
model alone is slightly more accurate than the fused forecast at every horizon.
In unemployment claims, the frozen recipe beats naive and the damped FA-ETS
reference, but not undamped FA-ETS, which is the better reference for that
series. In neither domain is there evidence that fusion beats the best single
model. The channel weights differ from housing, as the transfer hypothesis
expects: with the temporal channel that ignores seasonality, the seasonal
channel receives about three quarters of the weight at every horizon, and the
exchange of weight between the temporal and seasonal channels seen across
horizons in housing does not appear. Including the pandemic window raises
every method's MAPE by 0.7 to 2.2 points. Because an earlier pilot evaluation
scored claims data through August 2026, a clean holdout for this domain is not
available.

\begin{table}[t]
\centering
\caption{Unemployment insurance claims: test MAPE (\%) over six folds, pandemic
window excluded. The undamped FA-ETS row was added after the other results
were seen and is descriptive.}
\label{tab:ui_results}
\begin{tabular}{lccc}
\toprule
\textbf{Method} & 4 weeks & 8 weeks & 13 weeks \\
\midrule
Naive                        & 18.816 & 21.963 & 24.716 \\
Seasonally adjusted naive    & 17.836 & 19.294 & 20.613 \\
FA-ETS-d (damped)            & 17.868 & 19.411 & 20.791 \\
FA-ETS (undamped)            & \textbf{17.71} & \textbf{19.18} & \textbf{20.55} \\
FACTS-MAS (frozen recipe)    & 17.863 & 19.233 & 20.578 \\
\bottomrule
\end{tabular}
\end{table}

\subsection{Energy Forecasting}
\label{sec:energy}

Energy demand forecasting emphasizes short and medium cycles, weather
sensitivity, and abrupt regime changes driven by temperature extremes,
outages, and price shocks. The intended channel map for this domain is
Table~\ref{tab:energy_mapping}. No energy experiments are reported in this paper; the mapping below records
the planned configuration.

\begin{table}[t]
\centering
\caption{Planned FACTS-MAS evidence-channel mapping for the energy domain.}
\label{tab:energy_mapping}
\begin{tabular}{p{0.20\linewidth}p{0.70\linewidth}}
\toprule
\textbf{Channel} & \textbf{Energy-domain interpretation} \\
\midrule
Temporal
&
Historical energy-demand or load trajectory
\\
Exogenous
&
Weather, temperature, economic activity, price, or other external demand
drivers available at the forecast origin
\\
Seasonal
&
Recurring daily, weekly, and/or annual demand structure
\\
Intrinsic
&
Persistent characteristics of the forecasting region, grid, customer base,
or system capacity
\\
Event
&
Discrete weather, outage, policy, infrastructure, or other domain-relevant
events
\\
\bottomrule
\end{tabular}
\end{table}

The configuration is
$\mathcal{C}^{(\mathrm{energy})}=(\mathcal{E}^{(\mathrm{energy})}$,
$\mathcal{A}^{(\mathrm{energy})}$,
$\Theta^{(\mathrm{energy})}$,
$\mathcal{H}^{(\mathrm{energy})}$,
$\mathcal{R}^{(\mathrm{energy})})$.
Completed energy runs should report target, frequency, regions, coverage,
horizons, leakage controls, channel models, calibrated parameters, baselines,
and any regime definition, together with learned weights
$\mathbf{w}^{(\mathrm{energy})}=[w_{\mathrm{temporal}}$,
$w_{\mathrm{exogenous}}$,
$w_{\mathrm{seasonal}}$,
$w_{\mathrm{intrinsic}}$,
$w_{\mathrm{event}}]$,
or the subset of channels actually implemented. In this domain, seasonal and
exogenous channels are expected candidates for substantial weight when
validation supports that attribution.

\section{Discussion}
\label{sec:discussion}

The housing inventory case study shows that FACTS-MAS can decompose mixed evidence into
testable channels, tell those channels apart, calibrate domain assumptions,
and adapt their influence. It also clarifies what ``domain-agnostic'' does
and does not mean. The framework is not a pretrained model whose fitted
weights are expected to carry over unchanged to a new domain; instead, it carries
over a procedure: mapping evidence to functional channels and scoring each
channel against a reference, so as to calibrate, attribute, and fuse.

\subsection{Channel Diagnostics and Fusion Analysis}
\label{sec:analysis}

This section examines standalone channel value, domain-specific calibration,
and adaptive fusion: whether FACTS-MAS can distinguish useful evidence from
weak, redundant, or context-dependent evidence and adjust the fused forecast
accordingly.

\subsubsection{Evidence Channels as Testable Hypotheses}
\label{sec:testable_channels}

Each FACTS-MAS agent is a forecasting hypothesis: $\mathcal{H}_k$ asserts
that $E_k$ contains incremental information about future $y$. Including an
agent does not accept $\mathcal{H}_k$; historical validation decides whether
the channel beats the reference. Unlike a joint feature model, FACTS-MAS
produces independent agent forecasts before fusion, so weak evidence can be
diagnosed without removing or retraining unrelated channels. The housing
experiments show that pattern clearly. The temporal channel improves on persistence at 4 weeks but its advantage
narrows with horizon unless seasonality is removed first; seasonality helps at
every horizon; the tested macroeconomic magnitude channel adds little robust
incremental value; the intrinsic reversion calibrates to zero; and the initial
event-severity magnitude mapping is harmful before calibration. Agents are therefore candidate hypotheses, not mandatory
contributors.

\subsubsection{Macro Channel: A Well-Earned Null}
\label{sec:macro_null_analysis}

The Macro agent (exogenous channel) is driven by mortgage rates, monetary
policy, inflation, and unemployment. These are plausible housing drivers, but
plausibility does not establish short-horizon incremental forecast value.
Three diagnostics were used.

\paragraph{Direct forecast-signal validation.}
The agent's predicted modifier was compared with realized inventory changes
across the six backtest folds. The correlations were weak:

\begin{equation}
\begin{array}{c|ccc}
\text{Horizon}
&4\text{ wk}
&8\text{ wk}
&13\text{ wk}\\
\hline
r
&0.011
&-0.025
&-0.177
\\
p
&0.920
&0.814
&0.095
\end{array}
\end{equation}
providing no clear predictive signal.

\paragraph{Lag-sensitive analysis.}
A search over plausible transmission lags yielded apparently strong
correlations for several raw macroeconomic levels. Those correlations are
misleading: national macro values are identical across MSAs on a given date,
so MSA--date pairs pseudo-replicate one national signal and overstate the
effective sample size. Strong trends in both macro levels and inventory also
make simple level correlations vulnerable to spurious association.

\paragraph{Panel validation.}
The final diagnostic used a two-way fixed-effects panel with MSA and
year--quarter fixed effects, MSA-clustered standard errors, 5,850
observations, and the pre-specified midpoint lags rather than data-mined
best lags (Table~\ref{tab:macro_panel}).

\begin{table}[t]
\centering
\caption{Final housing Macro agent signal validation: $p$-values for the
original panel (MSA and year--quarter fixed effects, MSA-clustered errors),
with a Holm adjustment across the four variables, a wild cluster bootstrap
(Webb weights, 9,999 draws), and Driscoll--Kraay errors with MSA fixed effects.
$n=5{,}850$.}
\label{tab:macro_panel}
\begin{tabular}{lrrrrrr}
\toprule
\textbf{Variable} & \textbf{Lag} & \textbf{Coef.} & \textbf{Clustered} & \textbf{Holm} & \textbf{Wild bootstrap} & \textbf{Driscoll--Kraay} \\
\midrule
Mortgage rate & 6 weeks  &  0.0075 & 0.575 & 1.000 & 0.611 & 0.973 \\
Federal funds & 10 weeks & -0.0129 & 0.042 & 0.170 & 0.073 & 0.712 \\
CPI           & 10 weeks & -0.0047 & 0.085 & 0.254 & 0.086 & 0.463 \\
Unemployment  & 5 weeks  & -0.0015 & 0.548 & 1.000 & 0.577 & 0.264 \\
\bottomrule
\end{tabular}
\end{table}

Only federal funds crosses the conventional 5\% threshold under clustered
errors, and it does not survive the multiple-testing adjustment, the wild
bootstrap, or errors that allow for the common national regressor; mortgage
rates, the most direct theoretical housing-finance channel, are null under
every specification.
We treat this as a \emph{well-earned null}: the Macro agent was not retuned
until a positive result appeared, and its weak incremental skill is allowed
to reduce its fusion weight. FACTS-MAS can reject a plausible channel without
redesigning the architecture.

\subsubsection{Event Channel: Semantic Evidence and a Null Scale}
\label{sec:event_ablation}

The Event Agent fails differently. FEMA metadata contain useful semantic
information that an LLM can classify, but a qualitative severity score is not
automatically a valid continuous forecast adjustment. The original agent
returned an impact in $[-1,1]$. Prompt development showed that FEMA metadata
support only coarse severity tiers: similar assistance-program configurations
repeatedly mapped to the same severity, so the output behaved more like an
ordinal classifier than a continuous impact model. FACTS-MAS therefore tested
whether those magnitudes should modify the numerical forecast at all.

\paragraph{Impact-scale calibration.}
\label{sec:event_scale_analysis}

Let $m_t$ be the signed severity score and $\lambda$ the scale that maps it
into a forecast adjustment. Instead of assuming $\lambda>0$, calibration
tested $\lambda\in\{0,0.01,0.02,0.03,0.05,0.075,0.10\}$
(Table~\ref{tab:event_calibration}).

The ordering is monotonic: larger magnitude adjustments worsen validation
MAPE, and $\lambda=0$ is preferred in all six folds. With the scale at zero, the declaration-active window no longer affects the
forecast, and the event channel's output equals the naive reference. The
production Event Agent therefore keeps semantic classification and confidence
but removes the unsupported point-forecast magnitude adjustment. Useful
semantic information need not imply a validated numerical effect.

\subsubsection{From Raw Error to Softmax Weights}
\label{sec:weight_diagnosis}

\paragraph{Why raw accuracy is insufficient.}

An earlier fusion rule weighted agents by inverse validation MAPE. The Macro
agent, despite its null signal, sometimes received weight
comparable to stronger agents. A weakly fitted regularized model can stay close to
persistence; because persistence is competitive at short horizons, absolute
error can look good without adding new information. Absolute error therefore
confounds useful signal with naive imitation. FACTS-MAS uses skill relative
to naive, $s_{k,i}=1-e_{k,i}/e_{\mathrm{naive},i}$, so an agent that only
reproduces the naive forecast receives approximately zero skill even when its
MAPE looks competitive.

\paragraph{Robust skill aggregation.}
\label{sec:robust_skill}

Averaging pointwise relative-skill ratios is unstable when
$e_{\mathrm{naive},i}\rightarrow 0$: the ratio
$e_{k,i}/e_{\mathrm{naive},i}$ can explode. In one diagnostic, the temporal
agent had median pointwise skill about $+0.227$ but mean skill about
$-0.407$, with isolated ratios near $-32$. Fusion therefore uses
$\tilde{s}_k=\operatorname{median}_i(s_{k,i})$. The median protects
relative-skill aggregation whenever the reference occasionally has near-zero
error.

\paragraph{Why softmax replaced hard skill floors.}
\label{sec:softmax_analysis}

Transforming skill by $\max(s_k,\epsilon)^p$ avoided zero weights but
collapsed every negative-skill agent to the same floor, so $s=-0.01$ and
$s=-0.50$ became indistinguishable. Softmax,
$w_k=\exp(\tilde{s}_k/\tau)/\sum_j\exp(\tilde{s}_j/\tau)$,
preserves the full ranking, including among negative-skill agents. The fusion
rule therefore evolved from raw MAPE, to relative skill, to robust relative
skill with softmax, each step answering an observed failure of the previous
rule.

\subsubsection{Adaptive Fusion and Its Limits}
\label{sec:fusion_ablation}

\paragraph{Does regime conditioning carry information?}
\label{sec:regime_attribution}

Adaptive weighting chooses, for each fold and horizon, between pooled weights
and weights conditioned on regime and horizon. Its apparent 13-week benefit
was tested in three ways on the five-channel configuration, before the
temporal channel was replaced. First, with regime labels permuted in blocks,
46\% of 200 permutations at 8 weeks and 27\% at 13 weeks were at least as
accurate as the true labels. Second, weights conditioned on horizon alone,
with no regime, matched the regime-conditioned weights at 8 weeks (4.468\%
against 4.468\%) and were more accurate at 13 weeks (6.826\% against
6.987\%). Third, 60--75\% of the 13-week benefit came from a single fold, in
the stable regime. We therefore find no evidence that the regime labels carry
predictive information beyond horizon conditioning; with a national regime
variable and one substantive hiking episode in the sample, the regime
partition is close to a partition by historical period.

\paragraph{Data sufficiency.}
\label{sec:data_sufficiency}

Context-aware adaptation should depend on both estimated performance and
evidence volume. Let $n_{H,r}^{(f)}$ be the number of distinct validation
weeks for horizon $H$, regime $r$, and fold $f$. The production rule uses
regime-conditioned weights only if $n_{H,r}^{(f)}\geq 13$ and otherwise falls
back to the pooled weights. An earlier implementation had no such rule: when a
regime was absent from a validation window, every channel, including the
channels that reproduce the naive forecast, received a weight of 0.2. That
fallback affected 780 test cells, accounted for about one fifth of the
apparent 13-week benefit of regime weighting, and produced the apparent repair
of the cutting regime reported in an earlier version of this work.

\paragraph{Fusion versus the strongest single model.}
\label{sec:fusion_vs_single}

Once the temporal channel is seasonally adjusted, validation gives it 0.97 of
the weight at 8 weeks and 0.92 at 13 weeks, the seasonal channel receives most
of the remainder, and the exogenous channel receives almost none. The fused
forecast is then statistically equivalent to the factor-adjusted model alone
at 8 weeks in the backtest and on the holdout, and slightly less accurate at 4 and 13 weeks, in the backtest and on
the holdout. Three further candidates, each registered before it was run, did
not change this: a listing-flow channel built from Zillow new listings, newly
pending sales, and price-cut shares; fusion of deviations from the naive
forecast; and a validation-selected temperature with shrinkage toward equal
weights. None met the promotion rule, and the best was 0.002 percentage points
behind FA-ETS-d at 4 weeks ($p=0.42$). In the housing domain, the value of the
fusion step is therefore that it selects and recomposes the useful channels,
not that it combines them into a forecast more accurate than the best of them.

\subsubsection{Predictive versus Causal Attribution}
\label{sec:predictive_vs_causal}

Fusion weights measure predictive importance, not causation. For agent $k$,
$w_k$ is the influence assigned to its forecast given historical validation
skill; it does not claim that the underlying evidence caused the same share
of the target change. A low weight can mean little forecasting signal,
redundancy with another channel, a weak implementation, rare-regime
dependence, or insufficient validation history. Accordingly, \emph{factor
attribution} means predictive attribution within the forecasting
architecture, not structural causal identification.


\subsection{Why Full-Window Evaluation Matters}
\label{sec:housing_full_window}

The baseline experiments also exposed a significant evaluation artifact.
An earlier diagnostic evaluated one forecast origin per fold and horizon, and
its comparison of the temporal channel with the naive forecast scored
horizons that fell inside the embargo. Under that design the 13-week forecast
appeared to lose to naive in most MSAs. A single forecast date is a poor
estimator of general forecasting performance, so every result reported here
uses all eligible forecast origins in each test window: 2,610 MSA--origin
pairs per horizon in the backtest and 29 weekly origins per MSA in the holdout.

\subsection{What the Housing Results Imply}
\label{sec:discussion_housing}

FACTS-MAS improves on persistence at every evaluated horizon, on both Zillow
series, in the backtest and on the holdout, and it is more accurate than the
TimesFM foundation model. On the holdout a chain-of-thought language-model
forecast comes close to it, with no statistically significant difference.
It is not more accurate than the strongest single
model, factor-adjusted damped exponential smoothing, which it matches at 8
weeks. The more important architectural point is that the final forecast is
the product of tests that each channel had to pass. Temporal dynamics and
seasonality carry skill; the tested macro magnitude specification is largely
null; the intrinsic reversion and the event-magnitude map calibrate to zero.
That pattern is intentional. Available evidence is not the same as validated
predictive evidence: the first determines which agents can be proposed, and
the second determines how strongly they may move the forecast. A less
disciplined process could keep searching lags and prompts until a favorable
statistic appears; FACTS-MAS treats a null channel as an admissible outcome.

The Event Agent (event channel) and Macro Agent (exogenous channel) illustrate
two different limits of that distinction. An LLM can turn disaster metadata
into structured categories, explanations, and confidence, but that does not
imply a calibrated continuous effect; usefulness can stop at interpretation,
classification, ordinal severity, or only at a cardinal forecast effect. The
Macro agent fails for another reason: not semantic ambiguity, but too little
independent variation.
National macro series are repeated across MSAs, so naive correlations
overstate the sample, and a panel with thousands of entity--time rows can
still contain only a modest number of macro shocks.

Fusion weights should be read in the same spirit. For agent $k$, $w_k$ is a
predictive attribution, not a causal contribution: a large temporal weight
does not mean that temporal dynamics caused the same share of the observed
movement, and a small macro weight does not mean that monetary conditions are
causally irrelevant. Low weight can also mean redundancy, a weak
implementation, rare-regime dependence, or insufficient validation history.
Regime analysis shows why apparent context effects need a placebo. An earlier
version of the system seemed to fail in the hiking and cutting regimes and to
be repaired by regime-aware weighting, but the failure disappeared once the
seasonal channel was estimated properly, the repair was partly an artifact of
an equal-weight fallback, and randomly permuted regime labels did as well as
the true ones. Context-specific weighting reduces the observations behind
each decision, and the rare regimes where adaptation looks most attractive
often have the least support. The tradeoff is context specificity versus
estimation reliability. Energy or unemployment applications will meet the same
constraint whenever extreme weather or recessions are uncommon in the
validation sample.

\subsection{Toward Cross-Domain Transfer}
\label{sec:cross_domain}

Housing shows that the procedure works inside one heterogeneous problem.
It does not, by itself, show that the procedure is domain agnostic. Transfer
in FACTS-MAS is reconfiguration of evidence, not identity of fitted parts.

For source domain $\mathcal{D}_1$ and target domain $\mathcal{D}_2$, we do
not require $\mathcal{E}^{(\mathcal{D}_1)}=\mathcal{E}^{(\mathcal{D}_2)}$,
nor $\mathbf{w}^{(\mathcal{D}_1)}=\mathbf{w}^{(\mathcal{D}_2)}$.
Identical weights across unlike targets would be a warning that fusion is
insensitive to domain structure. What should be preserved is the procedure:
decompose the evidence, forecast from each channel, validate, calibrate,
attribute, and fuse. Formally, each domain supplies a configuration
$\mathcal{C}^{(\mathcal{D})}=(\mathcal{E}^{(\mathcal{D})}$,
$\mathcal{A}^{(\mathcal{D})}$,
$\Theta^{(\mathcal{D})}$,
$\mathcal{H}^{(\mathcal{D})}$,
$\mathcal{R}^{(\mathcal{D})})$,
where $\mathcal{E}$ is the evidence map, $\mathcal{A}$ the channel models,
$\Theta$ the calibrated parameters, $\mathcal{H}$ the horizons, and
$\mathcal{R}$ optional regimes. The invariant pieces remain the shared
interface, origin-time information boundary, leakage-safe rolling
evaluation, reference-relative skill, calibration, adaptive fusion, and
predictive attribution. Weights are re-estimated in the target domain:
$\mathbf{w}^{(\mathcal{D}_1)}\neq\mathbf{w}^{(\mathcal{D}_2)}$
is expected rather than treated as failed transfer. The transfer hypotheses
are therefore architectural, not parametric.

\paragraph{T1: Architectural transfer.}
A new problem can be written in the same functional channels and contracts
without a new orchestrator.

\paragraph{T2: Adaptive predictive value.}
Channel importance changes across domains and is learned from each domain's
own validation history.

\paragraph{T3: Forecasting benefit.}
After that reconfiguration, adaptive fusion should remain competitive with
the new domain's simple reference and strong standalone models.

The unemployment-claims study (Section~\ref{sec:unemployment}) supports T1:
the claims series was written in the same channels and contracts, and the
housing code ran unchanged. It supports T2: the learned weights differ from
housing, with the seasonal channel carrying most of the weight. It supports
T3 only in part: the fused forecast beats the naive reference and the damped
factor-adjusted model, but not the undamped factor-adjusted model, which
validation selected and which is the better single model for that series. The
energy configuration in Section~\ref{sec:energy} is planned and untested.

\section{Conclusion}
\label{sec:conclusion}

This paper introduced FACTS-MAS, a multi-agent forecasting framework that
treats heterogeneous evidence as a set of testable hypotheses rather than as
a single pooled feature list. Each functional channel produces an independent
forecast; leakage-safe validation measures incremental skill against a
reference; calibration is allowed to return a null effect; adaptive fusion
converts that evidence into weights that can depend on horizon and regime
when the sample can support the dependence.

On weekly U.S.\ housing inventory across 15 metropolitan areas, the frozen
configuration attains backtest MAPE of 2.118\%, 3.622\%, and 5.731\% at 4,
8, and 13 weeks on Zillow's smoothed series, against 4.194\%, 7.225\%, and
10.814\% for last-value persistence, and 1.243\%, 1.929\%, and 3.045\% on a
pre-registered holdout, against 3.122\%, 5.445\%, and 8.438\%. The gains are
smaller but still significant on the unsmoothed series. FACTS-MAS is more
accurate than the TimesFM~2.5 foundation model; on the holdout its error is
lower than that of a chain-of-thought language-model forecast built on the
NEXUS prompt, but not significantly so. It is not more accurate than a
factor-adjusted damped exponential smoothing model, with which it is
equivalent at 8 weeks. Those results coexist with null and rejected channels:
the tested macro magnitude specification adds little short-horizon skill,
validation removes the intrinsic reversion and the cardinal event-severity
adjustment, and regime-aware weighting shows no benefit beyond horizon
conditioning once a placebo is applied. The contribution of FACTS-MAS is
therefore the protocol itself: every channel is tested against a reference,
calibration may return a null, the final forecast is recomposed from the
channels that survive, and the result is checked on data held out until the
configuration is frozen. Transferring the method to a different domain
requires a new evidence map, new channel models where needed, and new
validation-based calibration and weights, but the FACTS-MAS architecture, the
shared interface, leakage-safe scoring, and adaptive fusion procedure remain
the same.

\bibliographystyle{plainnat}
\bibliography{references}

\end{document}